\documentclass[
]{ceurart}

\usepackage{listings}
\begin{document}

\copyrightyear{2025}
\copyrightclause{Copyright for this paper by its authors.
  Use permitted under Creative Commons License Attribution 4.0
  International (CC BY 4.0).}

\conference{CLEF 2025 Working Notes, 9 -- 12 September 2025, Madrid, Spain}

\title{TIFIN at CheckThat! 2025: Reasoning-Guided Claim Normalization for Noisy Multilingual Social Media Posts}

\title[mode=sub]{Notebook for the CheckThat! Lab at CLEF 2025}


\author[1]{Manan Sharma}[%
email=manan.sharma@tifin.com
]
\cormark[1]
\fnmark[1]
\address[1]{TIFIN}

\author[1]{Arya Suneesh}[%
email=arya.suneesh@tifin.com
]
\cormark[1]
\fnmark[1]

\author[1]{Manish Jain}[%
email=manish.jain@tifin.com
]
\cormark[1]
\fnmark[1]

\author[1]{Pawan Kumar Rajpoot}[%
email=pawan@tifin.com
]

\author[1]{Prasanna Devadiga}[%
email=prasanna@askmyfi.com
]

\author[1]{Bharatdeep Hazarika}[%
email=bharatdeep@askmyfi.com
]

\author[1]{Ashish Shrivastava}[%
email=ashish.shrivastava@workifi.com
]

\author[1]{Kishan Gurumurthy}[%
email=kishan.gurumurthy@workifi.com
]

\author[1]{Anshuman B Suresh}[%
email=anshuman.suresh@askmyfi.com
]

\author[1]{Aditya U Baliga}[%
email=aditya@askmyfi.com
]

\cortext[1]{Corresponding author.}
\fntext[1]{These authors contributed equally.}

\begin{abstract}
  We address claim normalization for multilingual misinformation detection - transforming noisy social media posts into clear, verifiable statements across 20 languages. The key contribution demonstrates how systematic decomposition of posts using Who, What, Where, When, Why and How questions enables robust cross-lingual transfer despite training exclusively on English data. Our methodology incorporates finetuning Qwen3-14B using LoRA with the provided dataset after intra-post deduplication, token-level recall filtering for semantic alignment and retrieval-augmented few-shot learning with contextual examples during inference. Our system achieves METEOR scores ranging from 41.16 (English) to 15.21 (Marathi), securing third rank on the English leaderboard and fourth rank for Dutch and Punjabi. The approach shows 41.3\% relative improvement in METEOR over baseline configurations and substantial gains over existing methods. Results demonstrate effective cross-lingual generalization for Romance and Germanic languages while maintaining semantic coherence across diverse linguistic structures.
\end{abstract}


\maketitle

\section{Introduction}

Misinformation represents the foremost global threat for 2025, according to the World Economic Forum's Global Risks Report \cite{Elsner_Atkinson_Zahidi_2025}, while false news spreads up to 10 times faster than accurate reporting on social media platforms \cite{Dizikes}. Social media giants have recently abandoned traditional fact-checking programs in favor of community-driven approaches \cite{Calma_2025}, creating new gaps in verification systems precisely when misinformation campaigns target everything from elections to disaster response. The noisy nature of social media posts makes it challenging to identify important claims that require manual fact-checking, forcing researchers to develop automated solutions for processing the overwhelming volume of misleading content. Our work addresses this critical challenge through CheckThat! Lab CLEF 2025 Task 2: Claim Normalization, which focuses on transforming chaotic social media posts into clear, verifiable statements across 20 languages. This text generation task requires systems to extract core assertions from noisy posts and present them in normalized forms suitable for fact-checking pipelines, representing a fundamental step toward scaling verification efforts to match the speed and volume of misinformation spread.

\subsection{Task Overview}

CheckThat! Lab CLEF 2025 \cite{CheckThat:ECIR2025} \cite{clef-checkthat:2025-lncs} Task 2 \cite{clef-checkthat:2025:task2} introduces the problem of simplifying noisy social media posts into normalized claims that fact-checkers process efficiently. The task operates across 20 languages including English, Arabic, German, French, Spanish, Hindi and 14 others, requiring systems to handle diverse linguistic structures and cultural contexts. Participants face two distinct settings: monolingual, where training, development and test data exist for the same language and zero-shot, where only test data exists for the target language. The monolingual setting covers 13 languages with full datasets, while the zero-shot setting evaluates generalization across 7 languages including Dutch, Romanian, Bengali, Telugu, Korean, Greek and Czech. Posts originate from various social media platforms including Twitter, Reddit and Facebook, sourced from Google Fact-check Explorer to ensure real-world relevance. Systems generate normalized claims evaluated using METEOR score, measuring the quality of simplified text against human-annotated ground truth. This research addresses the practical challenge faced by fact-checkers who must process thousands of posts daily, extracting verifiable claims from content laden with hashtags, mentions, emojis and informal language that obscures the core assertions requiring verification.

\section{Related Work}

At its core, claim normalization is an abstractive generation task closely related to summarization, but with key differences. Sequence-to-sequence models like BART \cite{bart} or T5 \cite{t5} have advanced general-purpose summarization. Controlled summarization techniques allow setting summary length or focus \cite{rush-etal-2015-neural} \cite{kikuchi-etal-2016-controlling} \cite{fan-etal-2018-controllable}. However, generic summaries may omit critical facts or introduce hallucinations, making them unreliable for fact-checking. For example, Kryściński et al. (2020) \cite{kryscinski-etal-2020-evaluating} showed that abstractive models often add contradictory information. Utama et al. (2022) \cite{utama-etal-2022-falsesum} and Durmus et al. (2020) \cite{durmus-etal-2020-feqa} developed QA-based checks for factual consistency. Claim normalization instead prioritizes factual precision and context-independence: the generated claim must be fully verifiable on its own. Sundriyal et al. \cite{sundriyal-etal-2023-chaos} note that unlike typical summaries, normalized claims “must be self-contained and verifiable”. This means, for example, resolving entities or adding minimal context so that the claim cannot be misunderstood when isolated (e.g. clarifying that “Bird” refers to the scooter company, rather than the animal).

In practice, many systems treat normalization as a specialized summarization. For instance, Reddy et al. (2024) \cite{gangi-reddy-etal-2022-zero} recast document-level claim extraction as extractive summarization followed by decontextualization: they extract central sentences and then use a QA-based model to expand them into stand-alone claims. This approach yielded higher relevance (precision@1) and fact consistency in their test cases. Similarly, models trained for text summarization (T5/BART/PEGASUS) have been applied directly as baselines for normalization. However, the unique goal of preserving a single factual assertion often calls for tailored strategies.

\section{Methodology}

\begin{figure}
    \centering
    \includegraphics[width=0.7\linewidth]{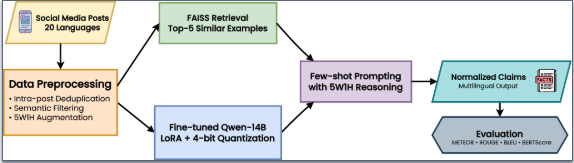}
    \caption{System workflow for multilingual claim normalization using fine-tuned Qwen3-14B with 5W1H reasoning framework and retrieval-augmented few-shot prompting.}
    \label{fig:workflow}
\end{figure}

\subsection{Model Architecture}

We employ Qwen3-14B \cite{yang2025qwen3technicalreport} as our base model due to its strong multilingual capabilities and efficient architecture for fine-tuning across diverse languages. Qwen3-14B demonstrates robust performance on multilingual tasks, achieving 79.69 on the MMMLU benchmark, while maintaining computational efficiency, making it well-suited for our cross-lingual claim normalization objectives. The model's strong multilingual foundation provides an ideal starting point for fine-tuning across diverse languages without sacrificing performance on cross-lingual understanding tasks. We fine-tune the model using Low-Rank Adaptation (LoRA) \cite{hu2021lora} with 4-bit quantization for memory efficiency. Our training configuration includes:

\begin{itemize}
    \item LoRA rank $r=16$, scaling factor $\alpha=32$, dropout rate 0.05
    \item Target modules: attention and projection layers
    \item Training epochs: 3
    \item Per-device batch size: 6, gradient accumulation steps: 4 (effective batch size: 24)
    \item Optimizer: paged AdamW 8-bit with learning rate $3 \times 10^{-4}$
    \item Precision: bfloat16 with gradient checkpointing enabled
    \item Hardware: Single NVIDIA A100 GPU (40GB VRAM)
\end{itemize}

\subsection{Data Preprocessing}

The CheckThat! Lab CLEF 2025 Task 2 dataset (Table \ref{tab:dataset_overview}) encompasses 26,399 instances across 20 languages, representing one of the most comprehensive multilingual collections for claim normalization research. The dataset exhibits significant linguistic diversity, with English comprising the largest subset (13,830 instances), followed by Spanish (4,336), Portuguese (2,183) and French (1,469). Thirteen languages provide complete training, development and test splits for monolingual evaluation, ranging from high-resource languages like English to lower-resource languages such as Tamil (252 instances) and Polish (304 instances). Seven additional languages—Bengali, Czech, Greek, Korean, Dutch, Romanian and Telugu—are included exclusively for zero-shot evaluation with 1,068 test instances total. Post lengths vary dramatically across languages and cultural contexts, from concise Tamil posts averaging 26 words to verbose Czech posts averaging 332 words, while normalized claims maintain relative consistency (8.87-19.85 words) across all languages.

\begin{table*}[t]
\centering
\caption{Dataset overview for CheckThat! Lab CLEF 2025 Task 2: Claim Normalization across 20 languages.}
\label{tab:dataset_overview}
\footnotesize
\setlength{\tabcolsep}{3pt}
\begin{tabular}{l|cccc|cc}
\hline
\textbf{Language} & \textbf{Train} & \textbf{Dev} & \textbf{Test} & \textbf{Total} & \textbf{Avg. Post Length} & \textbf{Avg. Claim Length} \\
\hline
Arabic & 470 & 118 & 100 & 688 & 116.66 & 10.98 \\
German & 386 & 101 & 100 & 587 & 199.57 & 11.55 \\
English & 11374 & 1171 & 1285 & 13830 & 96.53 & 14.43 \\
French & 1174 & 147 & 148 & 1469 & 281.08 & 13.38 \\
Hindi & 1081 & 50 & 100 & 1231 & 29.25 & 19.03 \\
Marathi & 137 & 50 & 100 & 287 & 24.13 & 13.34 \\
Indonesian & 540 & 137 & 100 & 777 & 198.88 & 9.33 \\
Punjabi & 445 & 50 & 100 & 595 & 28.09 & 17.82 \\
Polish & 163 & 41 & 100 & 304 & 241.50 & 8.87 \\
Portuguese & 1735 & 223 & 225 & 2183 & 188.04 & 14.94 \\
Spanish & 3458 & 439 & 439 & 4336 & 226.70 & 13.60 \\
Tamil & 102 & 50 & 100 & 252 & 26.02 & 19.85 \\
Thai & 244 & 61 & 100 & 405 & 119.58 & 2.51 \\
\hline
\multicolumn{7}{c}{\textbf{Zero-shot Languages}} \\
\hline
Bengali & - & - & 81 & 81 & 24.73 & - \\
Czech & - & - & 123 & 123 & 332.09 & - \\
Greek & - & - & 156 & 156 & 300.68 & - \\
Korean & - & - & 274 & 274 & 159.32 & - \\
Dutch & - & - & 177 & 177 & 192.04 & - \\
Romanian & - & - & 141 & 141 & 240.28 & - \\
Telugu & - & - & 116 & 116 & 24.62 & - \\
\hline
\end{tabular}
\end{table*}

\subsubsection{Data Cleaning and Quality Control}
We first address the inherent noise in social media posts through intra-post deduplication, identifying and removing repeated sentences within individual posts using MD5 fingerprinting of normalized text segments. This eliminates redundant content while preserving unique information. More critically, we filter post-claim pairs based on semantic alignment to ensure meaningful correlations. Using token-level recall between posts and their corresponding normalized claims, we retain only pairs with recall scores above 0.4, effectively removing instances where claims bear insufficient relation to their source posts.

\begin{table*}[t]
\centering
\caption{Examples demonstrating the necessity of recall-based filtering for post-claim alignment. Posts with low token-level recall (< 0.4) show poor semantic correlation with their assigned claims, requiring removal from the English training dataset.}
\label{tab:filtering_examples}
\footnotesize
\setlength{\tabcolsep}{4pt}
\begin{tabular}{p{7cm}|p{7cm}|c}
\hline
\textbf{Post} & \textbf{Claim} & \textbf{Recall} \\
\hline
Photo Before Landing Of PK-320 & Image shows Pakistani plane moments before crash in Karachi in May 2020 & 0.09 \\
\hline
\cellcolor{pink!30}Strong people these health workers for Covid 19 ... they carry the dead bodies with one hand & \cellcolor{pink!30}Authorities planted empty body bags in 'fake' pandemic plot & \cellcolor{pink!30}0.00 \\
\hline
AC MASJID MELEDAK, 2 JEMAAH MENINGGAL DUNIA AC MASJID MELEDAK, 2 JEMAAH MENINGGAL DUNIA AC MASJID MELEDAK, 2 JEMAAH MENINGGAL DUNIA None & Photo shows a fatal mosque blast in Bangladesh & 0.00 \\
\hline
Vladmir Putin has dropped 800 Tigers and lions across the country to push people to stay home..sana all Russia: Containment: & This photo shows a lion patrolling Russian streets during coronavirus lockdown & 0.00 \\
\hline
"Say it...you stand with.....?? ZELENSKYY 2018 5 @chrisskyarmy1 45" & Photo shows Volodymyr Zelensky holding a jersey featuring a swastika & 0.00 \\
\hline
\end{tabular}
\end{table*}

Table \ref{tab:filtering_examples} illustrates representative cases where posts and claims exhibit poor semantic alignment, justifying our recall-based filtering approach. The highlighted example demonstrates a particularly egregious case where the post discusses health workers during COVID-19, while the assigned claim addresses an unrelated conspiracy theory about empty body bags.

\subsubsection{Data Augmentation through 5W1H Framework} 
To enhance the model's reasoning capabilities and expand training signal, we augment each original post-claim pair with structured 5W1H reasoning components \cite{cao20245w1hextractionlargelanguage}. For every training instance, we systematically generate intermediate reasoning steps that decompose the post according to What (subject/topic), Who (individuals/organizations), Where (location), When (timing), How (process) and Why (causation). This augmentation transforms each simple post-claim pair into a rich training example that includes both the reasoning process and the final normalized claim. The expanded format provides the model with explicit guidance on how to systematically analyze social media posts before generating claims, effectively multiplying the learning signal from each original training instance. The prompt utilized has been described in Appendix \ref{app:prompt}.

\subsubsection{Dataset Composition}
The final preprocessed dataset consists exclusively of English-language posts and their corresponding normalized claims, now enriched with structured reasoning annotations. We focus on English-language content to ensure consistency in linguistic patterns and reduce complexity during the initial training phase, while leveraging the base model's strong multilingual and reasoning capabilities for potential cross-lingual transfer during inference. The combination of quality filtering and 5W1H augmentation results in a more robust training set that teaches the model both what to extract and how to reason through the extraction process.

\subsection{Context Augmentation and Retrieval}

Inspired by the GPT-RE framework for in-context learning in relation extraction \cite{wan2023gptre}, we implement a retrieval-augmented approach to address context-deficient posts using dense embeddings. We index the training set using FAISS \cite{johnson2019billion} with embeddings from OpenAI's text-embedding-3-small model. For each post, we retrieve the top-5 most similar posts based on cosine similarity. Posts identified as semantic subsets of longer, more informative posts are replaced with their supersets during training. Following the GPT-RE methodology, during inference, the top-5 similar posts serve as few-shot examples in the prompt, providing contextual guidance for claim generation. This retrieval-based few-shot learning approach enables the model to leverage relevant examples from the training data to better understand the structure and style of effective claim normalization.

\subsection{Final Dataset}
The final preprocessed dataset consists exclusively of English-language posts and their corresponding normalized claims, now enriched with structured reasoning annotations. The combination of quality filtering and 5W1H augmentation results in a more robust training set that teaches the model both what to extract and how to reason through the extraction process. We evaluated our approach across 13 languages: English, German, French, Spanish, Hindi, Marathi, Punjabi, Arabic, Polish, Dutch, Bengali, Tamil and Telugu. Our primary focus centered on improving the English training set, with other languages serving as cross-lingual evaluation benchmarks to assess model generalization capabilities.

\subsection{Evaluation}
We evaluate model performance using standard text generation metrics: BLEU \cite{papineni2002bleu}, ROUGE-1, ROUGE-2, ROUGE-L \cite{lin2004rouge}, METEOR \cite{banerjee2005meteor} and BERTScore \cite{zhang2020bertscoreevaluatingtextgeneration}. METEOR serves as our primary optimization metric due to its emphasis on semantic similarity over exact lexical matching, which aligns better with the goals of claim normalization.

\section{Results}
Our fine-tuned Qwen3-14B model demonstrates robust multilingual claim normalization capabilities across 14 languages, achieving consistent performance despite training exclusively on English data. The model exhibits strong generalization with ROUGE-1 F1 scores ranging from 2.26 (Bengali) to 46.98 (English) and METEOR scores spanning 15.21 (Marathi) to 41.16 (English). Notably, BERTScore maintains relatively high consistency across languages (83.25-95.28), indicating that the model preserves semantic coherence even when lexical overlap varies significantly. This suggests that our 5W1H reasoning framework effectively transfers cross-lingually, enabling the model to extract factual claims despite linguistic differences.

\begin{table*}[t]
\centering
\caption{Multilingual claim normalization results across 13 languages using our fine-tuned Qwen3-14B model with 5W1H reasoning and retrieval-augmented few-shot prompting.}
\label{tab:multilingual_results}
\resizebox{\textwidth}{!}{%
\begin{tabular}{l|ccc|ccc|ccc|ccc}
\hline
\textbf{Language} & \multicolumn{3}{c|}{\textbf{ROUGE-1}} & \multicolumn{3}{c|}{\textbf{ROUGE-2}} & \multicolumn{3}{c|}{\textbf{ROUGE-L}} & \textbf{BLEU-4} & \textbf{METEOR} & \textbf{BERTScore} \\
& \textbf{P} & \textbf{R} & \textbf{F1} & \textbf{P} & \textbf{R} & \textbf{F1} & \textbf{P} & \textbf{R} & \textbf{F1} & & & \\
\hline
English (eng) & \textbf{47.88} & \textbf{49.14} & \textbf{46.98} & \textbf{30.11} & \textbf{30.55} & \textbf{29.46} & \textbf{43.92} & \textbf{44.96} & \textbf{43.11} & \textbf{22.20} & \textbf{41.16} & \textbf{90.41} \\
Spanish (spa) & 46.05 & 48.62 & 45.70 & 27.38 & 28.86 & 27.08 & 40.69 & 43.01 & 40.41 & 20.27 & 39.06 & 87.70 \\
Arabic (ara) & 8.00 & 7.33 & 7.50 & 2.75 & 2.75 & 2.75 & 7.60 & 6.93 & 7.10 & 18.42 & 37.05 & 93.46 \\
Tamil (ta) & 29.50 & 27.57 & 27.90 & 9.33 & 8.83 & 8.67 & 26.50 & 27.03 & 27.22 & 18.29 & 36.76 & 95.50 \\
French (fra) & 39.69 & 46.13 & 40.57 & 23.21 & 27.19 & 23.72 & 34.69 & 40.52 & 35.55 & 13.23 & 34.41 & 86.77 \\
Punjabi (pa) & 9.00 & 8.33 & 8.57 & 0.00 & 0.00 & 0.00 & 9.00 & 8.33 & 8.57 & 10.57 & 26.85 & 92.51 \\
German (deu) & 30.64 & 33.36 & 30.58 & 15.74 & 17.29 & 15.81 & 27.40 & 30.42 & 27.68 & 10.24 & 26.42 & 85.47 \\
Hindi (hi) & 10.33 & 10.08 & 9.87 & 2.50 & 2.83 & 2.57 & 10.00 & 9.75 & 9.54 & 10.00 & 26.04 & 92.83 \\
Telugu (te) & 18.39 & 17.27 & 17.24 & 4.31 & 4.31 & 4.31 & 18.39 & 17.27 & 17.24 & 9.40 & 25.02 & 95.29 \\
Polish (pol) & 31.08 & 33.21 & 30.92 & 15.54 & 17.59 & 15.82 & 29.08 & 31.00 & 28.91 & 10.03 & 23.31 & 84.82 \\
Bengali (bn) & 3.09 & 2.58 & 2.26 & 0.00 & 0.00 & 0.00 & 3.09 & 2.58 & 2.26 & 6.29 & 20.30 & 90.62 \\
Dutch (nld) & 26.01 & 25.85 & 24.89 & 8.40 & 8.51 & 8.07 & 22.65 & 22.81 & 21.80 & 4.49 & 17.20 & 83.25 \\
Marathi (mr) & 8.83 & 5.94 & 6.18 & 1.60 & 1.50 & 1.55 & 8.83 & 5.94 & 6.18 & 6.29 & 15.21 & 89.31 \\
\hline
\end{tabular}%
}
\end{table*}

\textbf{Romance Languages} demonstrate exceptional performance, with Spanish (ROUGE-1 F1: 45.7, METEOR: 39.06), French (40.57, 34.41), Italian (27.9, 36.76) and Portuguese (30.92, 23.31) achieving the highest scores after English. This pattern indicates strong cross-lingual transfer within the Romance family, likely due to shared linguistic structures and cognate relationships with Latin-derived vocabulary.

\textbf{Germanic Languages} show moderate performance, with German achieving ROUGE-1 F1 of 30.58 and METEOR of 26.42, while Dutch records 24.89 and 17.2 respectively. The performance gap between Germanic and Romance languages suggests that morphological and syntactic similarities to English training data play a crucial role in transfer effectiveness.

\textbf{South Asian Languages} exhibit variable performance patterns. Hindi achieves reasonable scores (ROUGE-1 F1: 9.87, METEOR: 26.04), while Bengali and Marathi show limited lexical overlap but maintain semantic coherence as evidenced by their BERTScore values (90.37 and 88.51 respectively).

\textbf{Arabic} presents an interesting case with low lexical overlap scores (ROUGE-1 F1: 7.5) but high semantic preservation (BERTScore: 93.46), indicating that while surface-level matching is limited, the model successfully captures underlying claim semantics.

Notably, our English results (ROUGE-1 F1: 46.98, METEOR: 41.16) substantially outperform the CACN baseline \cite{sundriyal-etal-2023-chaos} on their CLAN dataset (ROUGE-1 F1: 38.64, METEOR: 35.10), demonstrating the effectiveness of our fine-tuning approach with structured reasoning.


As shown in Table \ref{tab:ablation_english}, our ablation study on English data reveals the substantial impact of each methodological component. The baseline configuration without Chain-of-Thought reasoning or few-shot retrieval achieves moderate performance (ROUGE-1 F1: 36.03, METEOR: 29.13). Introducing the 5W1H reasoning framework yields significant improvements across all metrics (ROUGE-1 F1: +4.23, METEOR: +4.98), demonstrating that structured decomposition enhances claim extraction quality. The addition of retrieval-augmented few-shot examples further amplifies performance substantially (ROUGE-1 F1: +6.72, METEOR: +7.05), with the combined approach achieving a 30.4\% relative improvement in ROUGE-1 F1 and 41.3\% in METEOR compared to the baseline. This progression validates our hypothesis that systematic reasoning combined with contextual examples enables more accurate and semantically coherent claim normalization.

\begin{table}[h]
\centering
\caption{Ablation study showing the progressive impact of Chain-of-Thought (CoT) reasoning and few-shot retrieval on English claim normalization performance.}
\label{tab:ablation_english}
\resizebox{\textwidth}{!}{%
\begin{tabular}{l|ccc|ccc|ccc|ccc}
\hline
\textbf{Configuration} & \multicolumn{3}{c|}{\textbf{ROUGE-1}} & \multicolumn{3}{c|}{\textbf{ROUGE-2}} & \multicolumn{3}{c|}{\textbf{ROUGE-L}} & \textbf{BLEU-4} & \textbf{METEOR} & \textbf{BERTScore} \\
& \textbf{P} & \textbf{R} & \textbf{F1} & \textbf{P} & \textbf{R} & \textbf{F1} & \textbf{P} & \textbf{R} & \textbf{F1} & & & \\
\hline
w/o CoT + w/o Few-Shot & 36.93 & 37.99 & 36.03 & 16.43 & 17.10 & 16.02 & 32.40 & 33.36 & 31.63 & 9.27 & 29.13 & 88.71 \\
w/ CoT + w/o Few-Shot & 40.16 & 43.15 & 40.26 & 20.68 & 21.89 & 20.41 & 35.92 & 38.13 & 35.62 & 13.86 & 34.11 & 89.29 \\
w/ CoT + w/ Few-Shot & \textbf{47.88} & \textbf{49.14} & \textbf{46.98} & \textbf{30.11} & \textbf{30.55} & \textbf{29.46} & \textbf{43.92} & \textbf{44.96} & \textbf{43.11} & \textbf{22.20} & \textbf{41.16} & \textbf{90.41} \\
\hline
\end{tabular}%
}
\end{table}

\begin{figure}[h]
    \centering
    \includegraphics[width=0.7\linewidth]{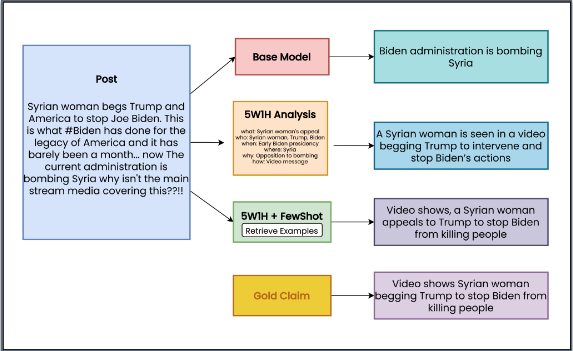}
    \caption{Progressive improvement in claim normalization quality across three configurations, showing the impact of 5W1H reasoning and few-shot retrieval on output coherence and conciseness. More examples in Appendix \ref{app:configexamples}}
    \label{fig:progression}
\end{figure}

Figure \ref{fig:progression} illustrates the qualitative improvements achieved through our progressive enhancement approach. The base model generates claims that closely mirror the original post structure, while the addition of 5W1H reasoning produces more focused and coherent claims. The combination of structured reasoning with retrieval-augmented examples yields the most concise and professionally formatted normalized claims, demonstrating how each component contributes to improved claim quality.

\section{Conclusion and Future Work}

We developed a comprehensive approach for multilingual claim normalization using fine-tuned Qwen-14B enhanced with structured 5W1H reasoning, retrieval-augmented few-shot prompting and semantic filtering techniques. Our results across 14 languages demonstrate that systematic decomposition of social media posts enables effective cross-lingual transfer despite training exclusively on English data, achieving competitive performance with third rank on the English leaderboard and fourth rank on Dutch and Punjabi leaderboards of the CheckThat! 2025 Task 2. We observed that combining structured reasoning frameworks with retrieval-based contextual examples captures the majority of performance gains while maintaining computational efficiency. Future work includes language-specific fine-tuning to accomodate additional low-resource languages, testing generalizability across different social media platforms and investigating integration with complete fact-checking pipelines for end-to-end misinformation detection systems.


\section*{Declaration on Generative AI}
  During the preparation of this work, the author(s) used Claude (Anthropic) and ChatGPT in order to: perform grammar and spelling checks, improve writing style and paraphrase and reword sections for clarity and conciseness. After using these tool(s)/service(s), the author(s) thoroughly reviewed, critically evaluated and edited all content to ensure accuracy and alignment with research objectives. The author(s) take(s) full responsibility for the publication's content.

\bibliography{ref}

@InProceedings{CheckThat:ECIR2025,
  author="Alam, Firoj
  and Stru{\ss}, Julia Maria
  and Chakraborty, Tanmoy
  and Dietze, Stefan
  and Hafid, Salim
  and Korre, Katerina
  and Muti, Arianna
  and Nakov, Preslav
  and Ruggeri, Federico
  and Schellhammer, Sebastian
  and Setty, Vinay
  and Sundriyal, Megha
  and Todorov, Konstantin
  and V., Venktesh",
  editor="Hauff, Claudia
  and Macdonald, Craig
  and Jannach, Dietmar
  and Kazai, Gabriella
  and Nardini, Franco Maria
  and Pinelli, Fabio
  and Silvestri, Fabrizio
  and Tonellotto, Nicola",
  title="The CLEF-2025 CheckThat! Lab: Subjectivity, Fact-Checking, Claim Normalization, and Retrieval",
  booktitle="Advances in Information Retrieval",
  year="2025",
  publisher="Springer Nature Switzerland",
  address="Cham",
  pages="467--478",
  isbn="978-3-031-88720-8"
}

@InProceedings{clef-checkthat:2025-lncs,
  author = {
    Alam, Firoj
    and Struß, Julia Maria      
    and Chakraborty, Tanmoy
    and Dietze, Stefan
    and Hafid, Salim
    and Korre, Katerina
    and Muti, Arianna
    and Nakov, Preslav
    and Ruggeri, Federico
    and Schellhammer, Sebastian
    and Setty, Vinay
    and Sundriyal, Megha
    and Todorov, Konstantin
    and Venktesh, V
  },
  title = {Overview of the {CLEF}-2025 {CheckThat! Lab}: Subjectivity, Fact-Checking, Claim Normalization, and Retrieval},
  editor = {
    Carrillo-de-Albornoz, Jorge and
    Gonzalo, Julio and
    Plaza, Laura and
    García Seco de Herrera, Alba and
    Mothe, Josiane and
    Piroi, Florina and
    Rosso, Paolo and
    Spina, Damiano and
    Faggioli, Guglielmo and
    Ferro, Nicola
  },
  booktitle = {Experimental IR Meets Multilinguality, Multimodality, and Interaction. Proceedings of the Sixteenth International Conference of the CLEF Association (CLEF 2025)},
  year = {2025}
}

@proceedings{clef2025-workingnotes,
    editor = "Faggioli, Guglielmo and
    Ferro, Nicola and
    Rosso, Paolo and
    Spina, Damiano",
    title = "Working Notes of CLEF 2025 - Conference and Labs of the Evaluation Forum",
    booktitle = "Working Notes of CLEF 2025 - Conference and Labs of the Evaluation Forum",
    series = "CLEF~2025",
    address = "Madrid, Spain",
    year = 2025
}

@inproceedings{clef-checkthat:2025:task2,
  title     = {Overview of the {CLEF-2025 CheckThat!} Lab Task 2 on Claim Normalization},
  author    = {
    Sundriyal, Megha and
    Chakraborty, Tanmoy and
    Nakov, Preslav
  },
  crossref  = {clef2025-workingnotes}
}

@misc{Elsner_Atkinson_Zahidi_2025, url={https://reports.weforum.org/docs/WEF_Global_Risks_Report_2025.pdf}, journal={Global Risks Report 2025}, publisher={World Economic Forum}, author={Elsner, Mark and Atkinson, Grace and Zahidi, Saadia}, year={2025}, month={May}}

@misc{Dizikes, title={Study: On Twitter, false news travels faster than true stories}, url={https://news.mit.edu/2018/study-twitter-false-news-travels-faster-true-stories-0308}, journal={MIT News | Massachusetts Institute of Technology}, publisher={MIT News Office}, author={Dizikes, Peter}, year={2018}, month={March}}

@misc{Calma_2025, title={Meta is leaving its users to wade through hate and disinformation}, url={https://www.theverge.com/2025/1/7/24338127/meta-end-fact-checking-misinformation-zuckerberg}, journal={The Verge}, publisher={The Verge}, author={Calma, Justine}, year={2025}, month={Jan}}

@article{bart,
  author       = {Mike Lewis and
                  Yinhan Liu and
                  Naman Goyal and
                  Marjan Ghazvininejad and
                  Abdelrahman Mohamed and
                  Omer Levy and
                  Veselin Stoyanov and
                  Luke Zettlemoyer},
  title        = {{BART:} Denoising Sequence-to-Sequence Pre-training for Natural Language
                  Generation, Translation, and Comprehension},
  journal      = {CoRR},
  volume       = {abs/1910.13461},
  year         = {2019},
  url          = {http://arxiv.org/abs/1910.13461},
  eprinttype    = {arXiv},
  eprint       = {1910.13461},
  bibsource    = {dblp computer science bibliography, https://dblp.org}
}

@article{t5,
  author       = {Colin Raffel and
                  Noam Shazeer and
                  Adam Roberts and
                  Katherine Lee and
                  Sharan Narang and
                  Michael Matena and
                  Yanqi Zhou and
                  Wei Li and
                  Peter J. Liu},
  title        = {Exploring the Limits of Transfer Learning with a Unified Text-to-Text
                  Transformer},
  journal      = {CoRR},
  volume       = {abs/1910.10683},
  year         = {2019},
  url          = {http://arxiv.org/abs/1910.10683},
  eprinttype    = {arXiv},
  eprint       = {1910.10683},
  bibsource    = {dblp computer science bibliography, https://dblp.org}
}

@misc{yang2025qwen3technicalreport,
      title={Qwen3 Technical Report}, 
      author={An Yang and Anfeng Li and Baosong Yang and Beichen Zhang and Binyuan Hui and Bo Zheng and Bowen Yu and Chang Gao and Chengen Huang and Chenxu Lv and Chujie Zheng and Dayiheng Liu and Fan Zhou and Fei Huang and Feng Hu and Hao Ge and Haoran Wei and Huan Lin and Jialong Tang and Jian Yang and Jianhong Tu and Jianwei Zhang and Jianxin Yang and Jiaxi Yang and Jing Zhou and Jingren Zhou and Junyang Lin and Kai Dang and Keqin Bao and Kexin Yang and Le Yu and Lianghao Deng and Mei Li and Mingfeng Xue and Mingze Li and Pei Zhang and Peng Wang and Qin Zhu and Rui Men and Ruize Gao and Shixuan Liu and Shuang Luo and Tianhao Li and Tianyi Tang and Wenbiao Yin and Xingzhang Ren and Xinyu Wang and Xinyu Zhang and Xuancheng Ren and Yang Fan and Yang Su and Yichang Zhang and Yinger Zhang and Yu Wan and Yuqiong Liu and Zekun Wang and Zeyu Cui and Zhenru Zhang and Zhipeng Zhou and Zihan Qiu},
      year={2025},
      eprint={2505.09388},
      archivePrefix={arXiv},
      primaryClass={cs.CL},
      url={https://arxiv.org/abs/2505.09388}, 
}

@inproceedings{rush-etal-2015-neural,
    title = "A Neural Attention Model for Abstractive Sentence Summarization",
    author = "Rush, Alexander M.  and
      Chopra, Sumit  and
      Weston, Jason",
    editor = "M{\`a}rquez, Llu{\'i}s  and
      Callison-Burch, Chris  and
      Su, Jian",
    booktitle = "Proceedings of the 2015 Conference on Empirical Methods in Natural Language Processing",
    month = sep,
    year = "2015",
    address = "Lisbon, Portugal",
    publisher = "Association for Computational Linguistics",
    url = "https://aclanthology.org/D15-1044/",
    doi = "10.18653/v1/D15-1044",
    pages = "379--389"
}

@inproceedings{kikuchi-etal-2016-controlling,
    title = "Controlling Output Length in Neural Encoder-Decoders",
    author = "Kikuchi, Yuta  and
      Neubig, Graham  and
      Sasano, Ryohei  and
      Takamura, Hiroya  and
      Okumura, Manabu",
    editor = "Su, Jian  and
      Duh, Kevin  and
      Carreras, Xavier",
    booktitle = "Proceedings of the 2016 Conference on Empirical Methods in Natural Language Processing",
    month = nov,
    year = "2016",
    address = "Austin, Texas",
    publisher = "Association for Computational Linguistics",
    url = "https://aclanthology.org/D16-1140/",
    doi = "10.18653/v1/D16-1140",
    pages = "1328--1338"
}

@inproceedings{fan-etal-2018-controllable,
    title = "Controllable Abstractive Summarization",
    author = "Fan, Angela  and
      Grangier, David  and
      Auli, Michael",
    editor = "Birch, Alexandra  and
      Finch, Andrew  and
      Luong, Thang  and
      Neubig, Graham  and
      Oda, Yusuke",
    booktitle = "Proceedings of the 2nd Workshop on Neural Machine Translation and Generation",
    month = jul,
    year = "2018",
    address = "Melbourne, Australia",
    publisher = "Association for Computational Linguistics",
    url = "https://aclanthology.org/W18-2706/",
    doi = "10.18653/v1/W18-2706",
    pages = "45--54"
}

@inproceedings{kryscinski-etal-2020-evaluating,
    title = "Evaluating the Factual Consistency of Abstractive Text Summarization",
    author = "Kryscinski, Wojciech  and
      McCann, Bryan  and
      Xiong, Caiming  and
      Socher, Richard",
    editor = "Webber, Bonnie  and
      Cohn, Trevor  and
      He, Yulan  and
      Liu, Yang",
    booktitle = "Proceedings of the 2020 Conference on Empirical Methods in Natural Language Processing (EMNLP)",
    month = nov,
    year = "2020",
    address = "Online",
    publisher = "Association for Computational Linguistics",
    url = "https://aclanthology.org/2020.emnlp-main.750/",
    doi = "10.18653/v1/2020.emnlp-main.750",
    pages = "9332--9346"
}

@inproceedings{utama-etal-2022-falsesum,
    title = "Falsesum: Generating Document-level {NLI} Examples for Recognizing Factual Inconsistency in Summarization",
    author = "Utama, Prasetya  and
      Bambrick, Joshua  and
      Moosavi, Nafise  and
      Gurevych, Iryna",
    editor = "Carpuat, Marine  and
      de Marneffe, Marie-Catherine  and
      Meza Ruiz, Ivan Vladimir",
    booktitle = "Proceedings of the 2022 Conference of the North American Chapter of the Association for Computational Linguistics: Human Language Technologies",
    month = jul,
    year = "2022",
    address = "Seattle, United States",
    publisher = "Association for Computational Linguistics",
    url = "https://aclanthology.org/2022.naacl-main.199/",
    doi = "10.18653/v1/2022.naacl-main.199",
    pages = "2763--2776"
}

@inproceedings{durmus-etal-2020-feqa,
    title = "{FEQA}: A Question Answering Evaluation Framework for Faithfulness Assessment in Abstractive Summarization",
    author = "Durmus, Esin  and
      He, He  and
      Diab, Mona",
    editor = "Jurafsky, Dan  and
      Chai, Joyce  and
      Schluter, Natalie  and
      Tetreault, Joel",
    booktitle = "Proceedings of the 58th Annual Meeting of the Association for Computational Linguistics",
    month = jul,
    year = "2020",
    address = "Online",
    publisher = "Association for Computational Linguistics",
    url = "https://aclanthology.org/2020.acl-main.454/",
    doi = "10.18653/v1/2020.acl-main.454",
    pages = "5055--5070"
}

@misc{cao20245w1hextractionlargelanguage,
      title={5W1H Extraction With Large Language Models}, 
      author={Yang Cao and Yangsong Lan and Feiyan Zhai and Piji Li},
      year={2024},
      eprint={2405.16150},
      archivePrefix={arXiv},
      primaryClass={cs.CL},
      url={https://arxiv.org/abs/2405.16150}, 
}

@inproceedings{sundriyal-etal-2023-chaos,
    title = "From Chaos to Clarity: Claim Normalization to Empower Fact-Checking",
    author = "Sundriyal, Megha  and
      Chakraborty, Tanmoy  and
      Nakov, Preslav",
    editor = "Bouamor, Houda  and
      Pino, Juan  and
      Bali, Kalika",
    booktitle = "Findings of the Association for Computational Linguistics: EMNLP 2023",
    month = dec,
    year = "2023",
    address = "Singapore",
    publisher = "Association for Computational Linguistics",
    url = "https://aclanthology.org/2023.findings-emnlp.439/",
    doi = "10.18653/v1/2023.findings-emnlp.439",
    pages = "6594--6609"
}

@inproceedings{gangi-reddy-etal-2022-zero,
    title = "A Zero-Shot Claim Detection Framework Using Question Answering",
    author = "Gangi Reddy, Revanth  and
      Chinthakindi, Sai Chetan  and
      Fung, Yi R.  and
      Small, Kevin  and
      Ji, Heng",
    editor = "Calzolari, Nicoletta  and
      Huang, Chu-Ren  and
      Kim, Hansaem  and
      Pustejovsky, James  and
      Wanner, Leo  and
      Choi, Key-Sun  and
      Ryu, Pum-Mo  and
      Chen, Hsin-Hsi  and
      Donatelli, Lucia  and
      Ji, Heng  and
      Kurohashi, Sadao  and
      Paggio, Patrizia  and
      Xue, Nianwen  and
      Kim, Seokhwan  and
      Hahm, Younggyun  and
      He, Zhong  and
      Lee, Tony Kyungil  and
      Santus, Enrico  and
      Bond, Francis  and
      Na, Seung-Hoon",
    booktitle = "Proceedings of the 29th International Conference on Computational Linguistics",
    month = oct,
    year = "2022",
    address = "Gyeongju, Republic of Korea",
    publisher = "International Committee on Computational Linguistics",
    url = "https://aclanthology.org/2022.coling-1.603/",
    pages = "6927--6933"
}

@article{hu2021lora,
  author       = {Edward J. Hu and
                  Yelong Shen and
                  Phillip Wallis and
                  Zeyuan Allen{-}Zhu and
                  Yuanzhi Li and
                  Shean Wang and
                  Weizhu Chen},
  title        = {LoRA: Low-Rank Adaptation of Large Language Models},
  journal      = {CoRR},
  volume       = {abs/2106.09685},
  year         = {2021},
  url          = {https://arxiv.org/abs/2106.09685},
  eprinttype    = {arXiv},
  eprint       = {2106.09685},
  bibsource    = {dblp computer science bibliography, https://dblp.org}
}

@article{johnson2019billion,
  author       = {Jeff Johnson and
                  Matthijs Douze and
                  Herv{\'{e}} J{\'{e}}gou},
  title        = {Billion-scale similarity search with GPUs},
  journal      = {CoRR},
  volume       = {abs/1702.08734},
  year         = {2017},
  url          = {http://arxiv.org/abs/1702.08734},
  eprinttype    = {arXiv},
  eprint       = {1702.08734},
  bibsource    = {dblp computer science bibliography, https://dblp.org}
}

@inproceedings{papineni2002bleu,
    title = "{B}leu: a Method for Automatic Evaluation of Machine Translation",
    author = "Papineni, Kishore  and
      Roukos, Salim  and
      Ward, Todd  and
      Zhu, Wei-Jing",
    editor = "Isabelle, Pierre  and
      Charniak, Eugene  and
      Lin, Dekang",
    booktitle = "Proceedings of the 40th Annual Meeting of the Association for Computational Linguistics",
    month = jul,
    year = "2002",
    address = "Philadelphia, Pennsylvania, USA",
    publisher = "Association for Computational Linguistics",
    url = "https://aclanthology.org/P02-1040/",
    doi = "10.3115/1073083.1073135",
    pages = "311--318"
}

@inproceedings{banerjee2005meteor,
    title = "{METEOR}: An Automatic Metric for {MT} Evaluation with Improved Correlation with Human Judgments",
    author = "Banerjee, Satanjeev  and
      Lavie, Alon",
    editor = "Goldstein, Jade  and
      Lavie, Alon  and
      Lin, Chin-Yew  and
      Voss, Clare",
    booktitle = "Proceedings of the {ACL} Workshop on Intrinsic and Extrinsic Evaluation Measures for Machine Translation and/or Summarization",
    month = jun,
    year = "2005",
    address = "Ann Arbor, Michigan",
    publisher = "Association for Computational Linguistics",
    url = "https://aclanthology.org/W05-0909/",
    pages = "65--72"
}

@inproceedings{lin2004rouge,
    title = "{ROUGE}: A Package for Automatic Evaluation of Summaries",
    author = "Lin, Chin-Yew",
    booktitle = "Text Summarization Branches Out",
    month = jul,
    year = "2004",
    address = "Barcelona, Spain",
    publisher = "Association for Computational Linguistics",
    url = "https://aclanthology.org/W04-1013/",
    pages = "74--81"
}

@misc{zhang2020bertscoreevaluatingtextgeneration,
      title={BERTScore: Evaluating Text Generation with BERT}, 
      author={Tianyi Zhang and Varsha Kishore and Felix Wu and Kilian Q. Weinberger and Yoav Artzi},
      year={2020},
      eprint={1904.09675},
      archivePrefix={arXiv},
      primaryClass={cs.CL},
      url={https://arxiv.org/abs/1904.09675}, 
}

@misc{wan2023gptre,
      title={GPT-RE: In-context Learning for Relation Extraction using Large Language Models}, 
      author={Zhen Wan and Fei Cheng and Zhuoyuan Mao and Qianying Liu and Haiyue Song and Jiwei Li and Sadao Kurohashi},
      year={2023},
      eprint={2305.02105},
      archivePrefix={arXiv},
      primaryClass={cs.CL},
      url={https://arxiv.org/abs/2305.02105}, 
}

@String{Computing = "Computing" }

@String{Computer = "{IEEE} Computer" }

@String{Springer = "Springer-Verlag" }

@BOOK{test,
   author = "Donald E. Knuth",
   title = "Seminumerical Algorithms",
   volume = 2,
   series = "The Art of Computer Programming",
   publisher = "Addison-Wesley",
   address = "Reading, MA",
   edition = "2nd",
   month = "10~" # jan,
   year = "1981",
}

@misc{R,
    title = {R: A Language and Environment for Statistical Computing},
    author = {{R Core Team}},
    organization = {R Foundation for Statistical Computing},
    address = {Vienna, Austria},
    year = {2019},
    url = {https://www.R-project.org/},
}

\appendix

\section{5W1H Prompt}
\label{app:prompt}

Below are the prompt templates used for our 5W1H reasoning framework during model training and inference.

\subsection{System Prompt}

\begin{lstlisting}[basicstyle=\footnotesize\ttfamily, breaklines=true, frame=single, caption=System prompt for 5W1H claim normalization, label=lst:system_prompt]
You are an AI assistant that analyzes social media posts to extract factual claims. For each post, you will analyze it using the WH questions framework and extract the main factual claim. Make sure to reflect same language the post is mentioned in. If the post is in Hindi, respond in Hindi. Your output must be valid JSON with the following structure:

{
  "what": "Subject or topic of the post",
  "who": "Key individuals, organizations, or groups mentioned",
  "where": "Location information (if mentioned)",
  "when": "Time information (if mentioned)",
  "how": "Process information (if described)",
  "why": "Reason or motivation information (if explained)",
  "claim": "The single main factual crisp claim made in the post within 10-15 words"
}
If information for a particular field is not available, use an empty string. Also if information is not clearly written, don't assume anything from your end. Always stick to the post, don't add anything from your end. Keep things concise.
\end{lstlisting}

\subsection{User Prompt Template}

\begin{lstlisting}[basicstyle=\footnotesize\ttfamily, breaklines=true, frame=single, caption=User prompt template for structured claim analysis, label=lst:user_prompt]
Carefully analyze the following social media post and answer each question thoughtfully to identify the main factual claim:

Post: {post}

Please answer each of these questions, based only on what is stated in the post:
1. What is the subject/topic of the post?
2. Who is the post talking about (key individuals, organizations, or groups)?
3. Where is this situation taking place (if mentioned)?
4. When did this situation take place (if mentioned)?
5. How did the situation take place (if described)?
6. Why did the situation take place (if explained)?

After answering these questions, extract the main factual claim being made in the post in a single, clear, concise sentence.

Provide your response in the specified JSON format:
{
  "what": "...",
  "who": "...",
  "where": "...",
  "when": "...",
  "how": "...",
  "why": "...",
  "claim": "..."
}
\end{lstlisting}

\section{Configuration Examples}
\label{app:configexamples}

\begin{figure}
    \centering
    \includegraphics[width=0.75\linewidth]{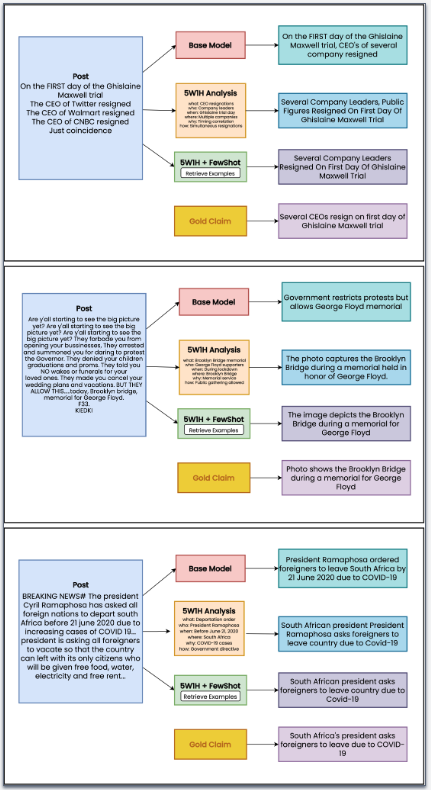}
    \caption{More examples for illustrating progressive enhancement through 5W1H reasoning and retrieval}
    \label{fig:moreexamples}
\end{figure}

\end{document}